\documentclass[journal,twoside]{IEEEtran} 

\usepackage{amsmath}
\usepackage{algorithmic}
\usepackage{array}%
\usepackage{fixltx2e}
\usepackage{stfloats}
\usepackage{url}
\usepackage{graphicx}
\usepackage{tabularx,ragged2e,booktabs,caption}
\usepackage[justification=centering]{caption}
\usepackage{tablefootnote}

\newcommand\T{\rule{0pt}{2.6ex}}       
\newcommand\B{\rule[-1.2ex]{0pt}{0pt}} 

\usepackage[T1]{fontenc}
\usepackage[utf8]{inputenc}

\usepackage{graphics} 
\usepackage{epsfig} 
\usepackage{mathptmx} 
\usepackage{times} 
\usepackage{amsmath} 
\usepackage{amssymb}  
\usepackage{amsfonts}
\usepackage{graphicx}
\usepackage{amsthm}
\usepackage{multicol}
\usepackage{array}
\usepackage{flushend}
\usepackage{cite}

\usepackage[table]{xcolor}    
\usepackage{color,colortbl}
\definecolor{Gray}{gray}{0.9}

\usepackage{pifont}
\usepackage{txfonts}
\usepackage{tikz} 
\newcommand*\circled[1]{\tikz[baseline=(char.base)]{
            \node[shape=circle,draw,inner sep=2pt] (char) {#1};}}
\usepackage{footnote}

\begin{document}

\title{Modular Relative Jacobian for Bipedal Robots}
\title{A Machine Learning Tool to Determine State\\of Mind and Emotion}

\author{Rodrigo~S.~Jamisola~Jr.
\thanks{R.S. Jamisola Jr. 
e-mail: {rjamisolajr@gmail.com,~jamisolar@biust.ac.bw}}
}

\vspace{0.5in}

\markboth{arXiv.org >> Computer Science >> Machine Learning}{Jamisola: A Machine Learning Tool to Determine State of Mind and Emotion}


\maketitle

\begin{abstract}
This paper investigates the possibility of creating a machine learning tool that automatically determines the state of mind and emotion of an individual through a questionnaire, without the aid of a human expert.
The state of mind and emotion is defined in this work as pertaining to preference, feelings, or opinion
that is not based on logic or reason.
It is the case when a person
gives out an answer to
start by saying,
"I feel...".
The tool is 
designed to mimic the expertise of a psychologist and
is built without any formal knowledge of psychology.
The idea is to build the expertise by purely computational methods
through thousands of questions collected from users. 
It is aimed towards possibly diagnosing
substance addiction, alcoholism, sexual attraction, 
HIV status, degree of commitment, activity inclination, etc.  
First, the paper presents the related literature and classifies them according to data gathering methods. 
Another classification is created according to preference, emotion, grouping, and rules to achieve a deeper interpretation and better understanding of the state of mind and emotion.
Second, the proposed tool is developed using an online addiction questionnaire with 10 questions
and 
292 respondents. In addition, an initial investigation on the dimension of addiction is presented 
through the built machine learning model.
Machine learning methods, namely, artificial neural network (ANN) and support vector machine (SVM),
are used to determine a true or false or degree of state of a respondent.
\end{abstract}


\IEEEpeerreviewmaketitle

\section{Introduction}
\label{sec:introduction}

{M}{achine} learning 
tools are not as widely used in psychology 
as in health and medicine. 
This is asserted by the fact that the interaction between cognition and emotion
is not yet fully understood \cite{Taylor2005313}.
Machine learning in medical applications
helped characterize 
genes and viruses 
\cite{
shipp2002diffuse,
guyon2002gene,
ye2003predicting,
shaik2014machine,
yang2015persistence};
\cite{magar2021potential,
shahid2021machine},
evaluate tumors and cancer cells
\cite{
dreiseitl2001comparison,
gletsos2003computer,
cruz2006applications,
kourou2015machine,
salgado2015evaluation,
ali2016computational};
\cite{myszczynska2020applications},
analyze medical images,
\cite{el2004similarity,
muller2004review,
salas2009analysis,
chaves2009svm,
greenspan2016guest,
macyszyn2016imaging};
\cite{willemink2020preparing},
and assess the health status of patients
\cite{
kononenko2001machine,
barakat2010intelligible,
o2012using,
prasad2016thyroid};
\cite{richens2020improving,
myszczynska2020applications}.
Other recent studies in machine learning include 
\cite{zheng2017multichannel,
chen2017brain,
hortensius2018perception,
jamone2018affordances,
cociu2018multimodal,
malete2019eegbased,
mmereki2021yolov3,
mohutsiwa2021eegbased
}.

A machine learning tool is proposed 
to mimic the expertise of a psychologist in determining the state of mind and emotion of an individual. 
In this paper, state of mind and emotion  
is referred to 
as something that is not based on conscious reasoning 
but is based on one's feeling or intuition. Thus normally, when we use it to give our preference or opinion, we start our statement by saying "I feel~...". 
The proposed method presented in this work does not claim any theoretical contribution to machine learning theory and is purely applied research. It investigates the possibility of duplicating the expertise of a psychologist through purely numerical computations, without any formal knowledge of psychology. 
It 
is based on an online questionnaire,
with inputs taken from 
online users and are analyzed using machine learning methods. 
Assessment questionnaires are extensively used 
in psychology and are analyzed 
by psychologists.

There are several advantages to the proposed approach. Firstly, the proposed tool can possibly replace the required expertise in performing an intelligent psychological evaluation. Secondly, a huge database of questions can be created such that a fresh set of questions can be provided for retakers. Thirdly, questions can be designed 
to be user-friendly in order to 
capture a fast response 
or to avoid a respondent from intentionally hiding a truthful answer. And lastly, through the proposed method, it is possible to determine the dimension of the state of mind and emotion by identifying critical questions that greatly influence the final output. It is noted that manual evaluation of this dimension can be very difficult to determine.  
This method
can possibly lead towards a deeper understanding of the state of mind and emotion, without the aid of a psychologist, but
through a wealth of questions and their classifications stored in a repository.


%

The author recognized the fact that 
questionnaire-based diagnosis cannot be very accurate
and precise. Issues on accuracy can occur 
because 
respondents can lie, and issues on precision can arise because 
even the respondent cannot
be precise about his or her own feelings
\cite{picard2001toward}.
(Subsequent references to ``his or her'' or ``he or she'' are omitted for simplification and are replaced by references to male sex only to refer to both sexes.)
However, the same challenges
are faced by questionnaire-based diagnostic examinations,
whether 
numerically or
professionally analyzed. 


There are other advantages of this numerical analysis compared to the human-analyzed questions. Firstly,
data errors in creating the model can be compensated by 
a statistically higher number of consistencies in the majority of the gathered information.
Secondly, 
analysis errors are consistent 
with the model and can be easily corrected by reconstructing the model. Compared to the manually analyzed questions, human error can contribute to errors in analysis.
And thirdly, updating the model can be fast by removing erroneous data, adding newly gathered data, and reconstructing the model.

It has long been suggested 
that machine learning models can provide better classification 
accuracy than explicit knowledge acquisition techniques \cite{ben1995classification}.
Thus in the past two decades, a
considerable number
of researches
were done in machine learning 
and has been applied to 
a wide range of fields of study. 
However, a more recent study by \cite{bollen2011modeling} showed that an analytic instrument from
empirical psychometric research can also
prove to be a valid 
alternative to machine learning to detect public sentiment.
In some cases, machine learning tools are used to solve traditional mathematical computations
\cite{jamisola2009using} 
which proved to be comparable to traditional results. Interestingly, the idea of a gaze sensor
that has the ability to detect staring, similar to that of humans was first discussed in \cite{jamisola2015oflove}.


To analyze the state of mind and emotion, 
there are two approaches used in this paper. The first is via a thorough discussion 
and analysis of 
related literature, and the second is via a machine learning model built on an addiction survey. 
In the first part of the paper, previous studies are classified according to data gathering methods
to establish the different modes of collecting information on choices by respondents that
are not based on reason or logic. 
This will introduce the reader to the wide range of mediums that the
information on the state of mind and emotion is collected, whether the respondent has
directly or indirectly provided the information. 
Then the type of choices is classified 
to look into their commonality and differences in order to gain a
deeper interpretation and better understanding of such choices.
The second part is dedicated to an initial attempt to build a machine learning model of addiction, which is identified
as a platform to investigate the state of mind and emotion. It also investigates
the dimension of addiction by verifying the independence or interdependence of the 
responses to the survey questions.

\begin{figure}[!t]
\centering
\includegraphics[width=0.5\columnwidth]{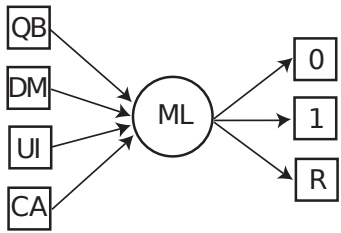} 
\caption{A diagram of the discussion presented in this paper. The center circle represents the machine learning (ML) method used for classification. Inputs discussed are questionnaire-based (QB), data mining (DM), user interface (UI), and camera (CA). The output of classification can be true (1), false (0), or a number range (R) representing an extent of influence.
}
\label{fig:diagram}
\end{figure}

From the extensive literature gathered, four data gathering methods
are identified namely, questionnaire-based, data mining, user
interface, and camera.
Fig.~\ref{fig:diagram} shows a diagram of Machine Learning (ML) discussion presented in this paper.  
Data-gathering methods are shown as blocks on the left-hand side: questionnaire-based (QB), data mining (DM), user interface (UI), and camera (CA). Possible outputs of machine learning analysis are true (1), false (0) or number range (R) indicating the 
extent of influence.
More recent studies in machine learning include
a review of probabilistic machine learning 
\cite{ghahramani2015probabilistic},
human-in-the-loop
\cite{holzinger2016interactive},
a review of recommender
systems \cite{adomavicius2015context},
and computational nature of
social systems \cite{hofman2017prediction}. 

Lastly, the following recent advances involve machine learning, questionnaire, or clinical assessment related to addiction. 
A 16-scale self-report questionnaire that assesses a range of addictive behaviors 
\cite{christo2003shorter}
uses traditional statistical analysis and does not use machine learning methods for classification.
A review paper \cite{huys2016computational} pointed out that computational psychiatry uses machine learning methods to improve disease classification, improve the selection of treatment or predict the outcome of treatment.
A study identifies risk factors using feature selection and predicts drinking patterns using cluster analysis
\cite{bi2013machine}. It used machine learning to classify from clinical data, but does not use online questionnaires or identify independent dimensions of addiction.
And, big data has been recognized unprecedented opportunity to track and analyze behavior
\cite{markowetz2014psycho}.

\section{Data-Gathering Methods}

This study proposes a questionnaire-based data-gathering method in building a machine learning model. 
To evaluate this method, 
different modes of data gathering in determining the state of mind and emotion
are presented here to show how they are used. 
These methods may present 
advantages as well as limitations
in output classification.
From the gathered literature, 
four types of data-gathering methods are identified:
questionnaire-based, online data mining, user interface, and camera. A quick-glance summary of 
the methods used is shown in Table~\ref{tab:data-gathering}.

In addition, the data gathering methods presented here 
may be classified into two types: one with direct user interaction, and another with indirect interaction. Direct user interaction includes user interface and camera. In this case, user response is immediately received
while interacting with the machine learning tool. This is normally done in real-time with at least one sensor involved. On the other hand, indirect user interaction includes questionnaire-based and online data mining, where the user response is saved and analyzed. Normally this is not done in real-time and no sensors are involved. This type of user interaction only requires regular office equipment and is thus cheaper to implement.


\subsection{Questionnaire-Based}

A questionnaire, also known as a survey, is used to gather information from respondents.
This is normally used to assess consumer satisfaction with products and services. 
One study used 
choice-based conjoint analysis that built models of consumer preferences
over products with answers gathered from questionnaires
\cite{chapelle2005machine,maldonado2017embedded,
huang2016consumer}.
This was a marketing research technique that was used to determine the required features of a new product 
based on feedback from consumers. Two machine learning tools were used: hierarchical Bayes analysis and Support Vector Machine (SVM). 

Another questionnaire-based study classified students for an intelligent
tutoring system in
an adaptive pre-test
using a machine learning tool 
\cite{aimeur2002clarisse}. Students were 
profiled
based on
performance measurements and gaming preferences through a questionnaire using
Bayesian network
and logistic regression
models
\cite{barata2016early}. 
In some cases, open-answer
questionnaires \cite{yamanishi2002mining}
were designed to use rule
learning and
correspondence
analysis
to 
automatically gather useful
information.
The authors argued that
answers to open-ended questions often contain valuable information and provide an important basis for business decisions.
This information included 
characteristics for individual analysis targets
and relationships among the targets.
A similar approach of information gathering was used
in
scoring open-ended responses to video prompts designed to assess Math teachers \cite{kersting2014automated}
using 
na\"{i}ve Bayes. 

Questionnaires to assess occupational exposure
\cite{wheeler2013inside} were used
to identify the underlying rules from 
responses through 
regression trees
and random forests.
In one study by \cite{terano1996knowledge}, the gathered data was used to
create efficient decision rules for extracting relevant
information from noisy questionnaire data. It
used both simulated breeding (genetic algorithm)
and inductive learning techniques.
Simulated breeding was used to get the effective features from the questionnaire data and inductive learning 
was used to acquire simple decision rules from the data. 
Through the use of questionnaires before and after deployment, one study \cite{karstoft2015early} predicted 
post-traumatic stress disorder of Danish soldiers after a deployment. It used a Markov boundary feature selection algorithm and classification used SVM.

\begin{table*}[h!]
\begin{minipage}{\textwidth}
\centering
\footnotesize
\caption[Caption for LOF]%
{Data-Gathering Methods used in Machine Learning Diagnostic Tools
\footnote{Abbreviated words meaning: 
net. -- network, 
cor. -- correspondence,
sim. -- simulated,
ind. -- inductive,
rand. -- random,
dec. -- decision,
n.n. -- nearest neighbor, and 
reg. -- regression.}}
\label{tab:data-gathering}      
\begin{tabular}{llll}
Purpose & Technique & Output & Reference  \T\B \\ 
\hline
\multicolumn{4}{l}{\T A. Questionnaire Based} \\ 
\hline \T
Conjoint analysis	   & Bayes, SVM     						& consumer feedback      & 				
										\cite{chapelle2005machine,maldonado2017embedded}\\
									& & & \cite{huang2016consumer}; 
									{\cite{oztas2021framework}}\\

Student classification & Bayesian net., SVM  					& categorized abilities & \cite{aimeur2002clarisse,barata2016early}\\
Analyze open answers   & rule learning, cor. analysis, na\"{i}ve Bayes & classification rules & \cite{yamanishi2002mining,kersting2014automated}\\ 
Extract relevant data & sim. breeding, ind. learning, rand. forests& decision rules	  		& \cite{wheeler2013inside,terano1996knowledge}\\ 
Assess traumatic disorder& Markov feature selection, SVM			& stress classification	& \cite{karstoft2015early}; 
{\cite{wani2021impact}}\\
%
\hline
\multicolumn{4}{l}{\T B. Online Data Mining} \\
\hline \T
Analyze sentiments 		  & na\"{i}ve Bayes, SVM, N-gram, dec. trees  	& sentiment classification			& \cite{Ye20096527,ortigosa2014sentiment}\\ 
Track behavior online 	  & clustering, neural network, Bayesian	  & dynamic user profile& 	
									\cite{smith2003mltutor,bidel2003statistical}\\
									&&&\cite{zhao2014personalized}\\
									
Identify FAQ or non-FAQ   & SVM, na\"{i}ve Bayes											& question classification	& \cite{razzaghi2016context}; 
{\cite{damani2020optimized}} \\
Assess review bias 			&  supervised ML, logistic regression 			& bias assessment & \cite{millard2016machine}; 
{\cite{didimo2020combining}} \\
Assess performance 			& PCA, neural network, SVM, dec. trees & credit score, marketing & \cite{koutanaei2015hybrid,moro2014data}\\
\hline 
\multicolumn{4}{l}{\T C. User Interface}\\ 
\hline \T
Detect human emotion  & ID3,  $k$-n.n., SVM, Bayesian net., reg. tree  	& emotion classification	& 			
										\cite{zacharatos2014automatic,lotfian2016practical}\\
										&&& \cite{chun2016determining,rani2006empirical}\\
										&&&\cite{chalfoun2006predicting,picard2001toward}\\
Infer user preference	& decision trees, HMM	& preference classification 					
& \cite{al2016predicting,bajoulvand2017analysis}\\
&&& \cite{chew2016aesthetic,cha2006learning}\\

Feedback to ML systems   & rule learning,  na\"{i}ve Bayes	& suggested features				& \cite{stumpf2009interacting}; 
{\cite{abid2020online}}\\
\hline
\multicolumn{4}{l}{\T D. Camera} \\
\hline\T\B
Detect real-time emotion 	& LDA, AdaBoost, SVM, Bayesian net.			& emotion classification		
& \cite{bartlett2005recognizing,littlewort2006dynamics}\\
&&&\cite{sebe2007authentic,shan2009facial}	\\

Video facial expression		& neurofuzzy, Markovian, na\"{i}ve Bayes			& emotion classification					
& \cite{ioannou2005emotion,cohen2003facial}\\
3D facial expression 			&  LDA					& 3D facial	database				
&\cite{yin20063d}; 
{\cite{lin2020orthogonalization}}\\
\hline

\end{tabular}

\end{minipage}
\end{table*}

\subsection{Online Data Mining}

Online data mining involved an automatic gathering of information from online content.
This is normally performed by applications that crawl through them and 
gather information based on keywords found. The mined data may
carry information
about personal sentiments, opinions, or preferences. This data was also used to track user's behavior online and assess a person's response or performance. 

On sentiment analysis, one study \cite{Ye20096527} considered
sentiment classification 
of online reviews as a class of web-mining techniques that performed an analysis of opinion 
on travel destinations.
The authors gathered information 
from travel blogs, then used three supervised machine learning 
techniques, namely, na\"{i}ve Bayes, SVM, and the character-based N-gram mode to come up with sentiment
classification.
Another study analyzed sentiments on Facebook 
comments \cite{ortigosa2014sentiment}, and used 
decision trees, 
na\"{i}ve Bayes,
and 
SVM to classify them.
A user's behavior online was tracked through
a user's browsing history in hypertext \cite{smith2003mltutor}.
This study involved 
applying machine learning algorithms to generate personalized adaptation 
of hypertext systems. 
Conceptual clustering 
and inductive machine learning algorithms were used.
Predefined user profiles were replaced with a dynamic user profile-building scheme in order to provide individual adaptation. 
A homemade access log database
was used, together with a number of statistical
machine learning models, to compare different classification or 
tracking of user
navigation patterns for closed world hypermedia
\cite{bidel2003statistical}.
Neural network and Markovian models were used
in dealing with temporal data.
Another study exploited the rich user-generated location contents in location-based social
networks \cite{zhao2014personalized}
to offer tourists the most relevant and personalized local venue recommendations using 
the Bayesian approach.

In searching for online help, users may ask
questions that can be frequently asked questions
(FAQ) or not. A study identified FAQs from non-FAQs
\cite{razzaghi2016context} by using
machine learning-based parsing and question classification.
The authors noted that the identification of specific question features was the key to obtaining an accurate FAQ classifier. The SVM method and na\"{i}ve Bayes were used.
Risk-of-bias assessment can be a very critical
factor in systematic reviews.
One study tackled this issue \cite{millard2016machine}
and created three risk-of-bias assessment properties: sequence generation, allocation concealment, and blinding. The approach used supervised machine learning
and logistic regression.

Online data mining also considered performance assessment in classifying credit scores and telemarketing success. 
In classifying credit scores,
one study \cite{koutanaei2015hybrid}
used feature selection algorithms and
ensemble classifiers that include
principal component analysis,
genetic algorithm, artificial neural network,
and AdaBoost.
Telemarketing success 
\cite{moro2014data}
predicted the success of telemarketing calls for selling bank long-term
deposits using
logistic regression, decision trees, neural networks, and SVM.

\subsection{User Interface}

Previous studies in
user interface data gathering can be classified into two types: one through user input
and another is by detection of brain signals. The first type of user interface
allowed the respondent to input his 
reaction to a stimulus,
normally through a screen display, 
by natural language,
or physical cues. The second type 
is through the detection of brain signals, normally through
electroencephalogram (EEG), which
allowed real-time interaction
with the respondent and online machine learning analysis.  
It can be used to detect human emotion, infer user preference, and as a
feedback mechanism to machine learning systems.

Emotion recognition from body movements 
\cite{zacharatos2014automatic}
used cameras and motion tracker sensors to track body
movements.
To classify the emotion of the user, the study
used Principal Component Analysis (PCA), na\"{i}ve Bayes, and Markov model. 
Emotion classification by speech \cite{lotfian2016practical} was studied where  
%
a speech emotion retrieval system aimed to detect a subset of data with specific expressive content.
The experiment used a speech sensor to collect data and SVM for emotional classification.
A study used many types of sensors, including temperature sensors to detect ambient temperature, and
to collect data 
\cite{chun2016determining} in
a patent application to determine emotion.
Data acquisition devices include a camera, a microphone, an accelerometer, a gyroscope, a location sensor, and a temperature sensor to detect ambient temperature.
This study outputted emotion classification using PCA and SVM.

A study was performed to detect human emotion from physiological cues
using four machine learning
methods: $k$-nearest neighbor, regression tree,
Bayesian network, and SVM \cite{rani2006empirical}.
The respondents interact with computers, and their
emotions were detected by sensors attached to their bodies.
Results showed that SVM gave
the best classification accuracy even though all the
methods performed competitively.
ID3 (Iterative Dichotomiser 3) algorithm is used in 
machine learning and natural language processing domains. 
For the study in \cite{chalfoun2006predicting}, 
the learner's emotional reaction in a distant learning environment
is inferred using the ID3 algorithm. 
A study by 
\cite{picard2001toward}
used physiological signals 
to gather data from a single subject over six weeks.
A computer-controlled prompting system called ``Sentograph''
showed a set of personally-significant imagery
to help elicit eight emotional states, namely,
no emotion (neutral),
anger, hate, grief, platonic love, 
romantic love, joy, and reverence.
Transforming techniques used 
sequential floating
forward search, Fisher projection, and
a hybrid of the two.
Classifiers used $k$-nearest-neighbor
and maximum a posteriori.
One study \cite{wall2015mapping} considered
a prediction of emotional perceptive competency and
implicit affective preferences. It gathered data through eye-tracking and neurocognitive processes
comprising of six domains:  executive function and attention,
language, memory and learning, sensorimotor, visuospatial
processing, and social perception. They were used to predict emotion through
linear regression, PCA
with linear regression, and 
SVM. 

The study in \cite{al2016predicting}
considered inferring interface design preferences from the user’s eye-movement behavior
using an eye tracker device. Machine learning 
information processing is done via decision trees,
and this outputs user design preferences.
Folk music preference was studied in
\cite{bajoulvand2017analysis} using EEG signals to collect data from the user. 
SVM classifier with radial basis function (RBF) kernel was used, and the output
is a predicted user preference.
Another study in user preference was \cite{chew2016aesthetic}
that considered aesthetic preference recognition of 3D shapes. 
It gathered user information through EEG signals and used
SVM and k-nearest neighbors to process the information. 
A user interface has been devised so different learner preferences can be 
acquired through interaction with the system \cite{cha2006learning}. 
Based on this information, user interfaces were customized to accommodate a learner's 
preference in an intelligent learning environment. 
User preference was diagnosed using decision tree and hidden Markov model (HMM) approaches.

One study used respondents to communicate feedback to machine learning systems 
\cite{stumpf2009interacting},
with the purpose of improving its accuracy. 
Users were shown explanations of machine learning predictions and were asked to provide feedback.
These include suggestions for re-weighting of features, proposals for new features, feature combinations, 
relational features, and changes to the learning algorithm.
Two learning algorithms were used: the Ripper rule-learning algorithm and the
na\"{i}ve Bayes algorithm.
The study showed the potential of rich human-computer collaboration via on-the-spot interactions,
to share intelligence between user and machine.

\subsection{Camera}

The last method discussed in the paper for data gathering 
is through the use of a camera.
This method can perform a real-time observation
of 
facial expression, or can be non-real-time 
through a video recording,
which the machine learning
method then analyzed to output a judgment. 
One 
disadvantage of relying on face or voice to judge a person's emotion
is that we may see a person smiling or hear that her voice sounded cheerful,
but this does not mean that she was happy \cite{picard2001toward}.
But because human emotion is greatly displayed by facial expression, 
its detection by a camera is extensively studied.


To predict negative emotion, one study \cite{hung2016predicting}
made use of the mobile phone camera
and processed the information using a na\"{i}ve Bayes classifier, decision tree, and 
SVM. 
A study that used a camera to detect facial expression
\cite{bartlett2005recognizing,littlewort2006dynamics} utilized 
AdaBoost for feature selection prior to classification by SVM 
or linear discriminant analysis (LDA).
Facial expressions in the video were analyzed in a study in \cite{sebe2007authentic}.
It developed an authentic facial expression database where 
the subjects showed natural facial expressions based on their emotional state. 
Then it evaluated machine learning algorithms for emotion detection 
including Bayesian networks, SVMs, and decision trees.
Local binary pattern (LBP) was used for facial expression recognition. 
The authors used
boosted-LBP to extract the most discriminant LBP features, 
and the results were classified via SVM.
It was claimed that the method worked in low resolutions of face images
and compressed low-resolution video sequences captured in real-world environments
\cite{shan2009facial}.

Extraction of appropriate facial features 
and identification of the user's emotional state 
through the use of a neurofuzzy system was studied
\cite{ioannou2005emotion},
which can be robust to variations among different people.
Facial animation parameters 
are extracted from ISO MPEG-4 video standard. 
The neurofuzzy analysis was performed based on the rules
from facial animation parameters variations 
both at the discrete emotional space
and 2D continuous activation–evaluation. 
The multi-level architecture of 
a hidden Markov model layer 
was shown in \cite{cohen2003facial} 
for automatically segmenting 
and recognizing human facial expressions 
from video sequences. 
Classification of  expressions from video
used na\"{i}ve Bayes classifiers
and learning the dependencies among different facial motion features used Gaussian tree-augmented
na\"{i}ve Bayes classifiers.

A 3D facial expression recognition was shown in 
\cite{yin20063d} that has developed 3D facial expression database.
It has created 
prototypical 3D facial expression shapes and 2D facial textures of 2,500 models from 100 subjects. 
LDA
classifier was used to classify the prototypic facial expressions of
sixty subjects.	
From all the four types of data-gathering methods, one may say that the sensor-based inputs with direct user interaction can be considered to be more accurate
compared to the indirect, non-sensor-based method where the user inputs
may be subjective.
However, sensor-based inputs may be 
limited to what a sensor can detect. 
For example, a camera may detect a smiling face, but it does not mean the person is happy. 
Or if the sensor is not accurate enough to detect the brain signal, it can give out other output instead of what the user intended. 
Thus if the user is objective in his inputs, the questionnaire-based or data-mining method may be more accurate than the sensor-based. 
Until the time that more sophisticated sensors are developed to detect accurately what the person really wants to convey, 
at the current technological state,
the questionnaire-based method may
significantly cover the user's actual
state of mind and emotion. 

\section{Types of Classifications}


From the previous section, we were able to analyze the method of 
data gathering in the state of mind and emotion, including 
the questionnaire-based method that is proposed in this work. 
Using the same literature discussed in the previous section, 
together with a few more additional pieces of literature, we propose four ways
in classifying the state of mind and emotion once data has been gathered.
The proposed four classifications are
preference, emotion, grouping, and rules.

\subsection{Correlation Among Classifications}

In this work, preference is referred to as an individual's intuitive choice 
given two or more options. It does not include any emotion.
For visualization purposes, preference can be thought of as a 
``horizontal'' expression of one's feelings,
where the emotional level remains ``flat.''
On the other hand, emotion is not a 
conscious
choice but an individual's reaction to an outside stimulus that affects the person's disposition. 
Emotion is not based on intuition
because intuition involves a 
mental
process 
without conscious reasoning. 
It can be thought of as a
``vertical'' 
expression of feelings 
with varying intensity. Thus 
the usual reference to ``up-and-downs'' of emotion.
Preference and emotion are direct results from individual responses and are normally referred to as feelings.
In other words, preference is a ``non-emotional'' feeling and is intuitive, while emotion is an ``emotional'' feeling and is not intuitive. Therefore feelings involve a mental process (intuition) and a non-mental process (emotion).

On the other hand, grouping and rules classification are not direct results from individual responses.
Rather, 
individual responses are further analyzed to output a final judgment. In grouping, classification rules are applied to the individual responses to classify them according to a set grouping. In rules, the individual responses are used to create new rules or modify existing ones, which may later be used to arrive at a final judgment. 
In terms of the interaction with the respondents, classifications by rules and 
grouping normally may entail an indirect interaction, while the
emotion and preference classifications normally require direct interaction and are usually performed in real-time. 
And lastly, in terms of decision outcomes, 
emotion and preference classification are normally decided by the user. In grouping and rules classification, the decision outcomes are normally decided by an observer. Table~\ref{tab:classification} shows the summary of classifications.

\begin{table}[tb!]
\footnotesize
\caption{Classifications in Determining\\State of Mind and Emotion}
\label{tab:classification}
\begin{tabularx}{\columnwidth}{ll}
\multicolumn{1}{c}{Purpose} & \multicolumn{1}{c}{References}  \\ 
\hline \T\B

A. Classification: Preference & 	\\
- Consumer product & \cite{toubia2007optimization,chapelle2005machine,maldonado2017embedded}\\

- Travel destination interest & \cite{Ye20096527,zhao2014personalized,li2021machine}\\
- Ranking aesthetic preferences & \cite{Hullermeier20081897,al2016predicting,bajoulvand2017analysis}\\
 
- Analyze online sentiments & \cite{cha2006learning,stumpf2009interacting,budhi2021using}\\
- Tracking of navigation patterns & \cite{smith2003mltutor,bidel2003statistical,kumar2020progressive}\\
\hline\T\B

B. Classification: Emotion & \\ 	
- Detect emotion from physiological cues & \cite{rani2006empirical,picard2001toward,chun2016determining}\\
										
- Emotion detection from speech & \cite{Devillers2005407,freitag2000machine,agrawal2020speech}\\
- Online facial expression from camera & \cite{bartlett2005recognizing,littlewort2006dynamics,fathima2020review}\\
										
- Off-line facial expression from video & \cite{sebe2007authentic,cohen2003facial,jin2020diagnosing}\\

\hline\T\B

C. Classification: Grouping & \\	
- Student abilities prediction from response  & \cite{beck2000high,aimeur2002clarisse,lamb2021computational}\\
											 
- Identifying off-task behavior & \cite{cetintas2010automatic,karstoft2015early,abou2020application}\\
								
- Model formation to predict future actions & \cite{webb2001machine,razzaghi2016context}\\
- Gaming-detection model for tutoring behavior  & \cite{walonoski2006detection,barata2016early}\\			
			
\hline\T\B

D. Classification: Rules & \\ 		
- Open answers to questionnaires  & \cite{yamanishi2002mining,kersting2014automated}\\
- Efficient decision rules from noisy data & \cite{terano1996knowledge,wheeler2013inside,rolf2020balancing}\\
- Learning casual relationships, word meanings & \cite{Boose1985495,Tenenbaum2006309,huang2020detecting}\\
- Production rules from independent assessment & \cite{jarvis2004applying,millard2016machine}\\
\hline

\end{tabularx}
\end{table}

\subsection{Classification by Preference}

Preference is an option chosen by an individual based on how he 
feels, but with no emotion attached to the judgment.
It is mostly
used to identify the liking of a user to a particular 
person, place, product, or service.
Traditionally, this method of gathering preference information from users was used by many companies 
\cite{toubia2007optimization,chapelle2005machine,maldonado2017embedded,huang2016consumer}
to assess their current market share or to estimate the degree of acceptance of a new product introduced to the market.

In recent years, user preference posted online is becoming a new and powerful approach in gathering and analyzing such information.
One approach was by tracking navigation patterns online and present the most likely information that will be of interest to the user based on navigation preference 
\cite{smith2003mltutor,bidel2003statistical}.
It can be used to present the most likely advertisements, interactive interfaces, or locations of places that will be of interest to the user. 
Another approach was tracking user preference on travel destinations 
\cite{Ye20096527,zhao2014personalized} or
preference on aesthetics  
\cite{Hullermeier20081897,al2016predicting,bajoulvand2017analysis,chew2016aesthetic}
or of online sentiments
\cite{cha2006learning,ortigosa2014sentiment,stumpf2009interacting}.

One can say that such information is monetarily driven by companies providing products and services. But this can also be very helpful to users who might want to find immediate solutions to urgent needs. Thus nowadays, matching demand to its solution can be quite easily performed by analyzing online user preference.
The other advantage of posting preference online is that the online document can become the source of information for other users. For example, other users can put additional online reviews for a particular travel destination, making the expanded information more exhaustive and useful for potential visitors. This is also true for political sentiments that gather huge support given a short period of time. This has been a vehicle of many social actions within the past decade.
Thus online preference can become a powerful tool for users of the same liking. This enables them to bargain for  better service or initiate a desired social change.

\subsection{Classification by Emotion}

Of the four types, emotion classification is quite extensively studied. It is normally detected through camera or EEG signals, and 
it enjoys significant interest among researchers.
Emotion is an 
expression of the feelings of an individual with varying intensity according to the degree of feelings conveyed. 

Emotion can be transmitted and be strongly shared among individuals, as in a mob. It has the ability to overpower all other senses of the individual to assume singularity of purpose.
The tone of emotion can be set given an appropriate environment, as in a relaxing environment with soft music and dim lights. Or it can be instantaneously derived by giving the right stimulus as in the case of anger by striking a sensitive chord or happiness by watching a cute baby.
Normally, there are three types of emotional stimulus, namely, visual, by hearing, and by touching. Thus a person can be stimulated visually as in a movie that is horror, comedy, or sexually explicit; or stimulated by sound as in a vile language or shouting or relaxing music; or stimulated by touching as in shaking hands, hugging or
kissing, or strong dislike by hard grip or punching.

As emotion can be stimulated visually, or by sound, or by touching, it can also be manifested in the same manner.
Thus from previous studies, visual manifestation of emotion 
through facial expression was identified in real-time using camera
\cite{bartlett2005recognizing,littlewort2006dynamics,ioannou2005emotion,yin20063d}
or off-line using recorded video
\cite{sebe2007authentic,cohen2003facial,shan2009facial}.
Emotion was also identified by sound through speech \cite{Devillers2005407,freitag2000machine,lotfian2016practical}.
The last method may not be obvious from a human observer when emotion was detected through the use of sensors attached to the body, and was identified from bodily cues \cite{rani2006empirical,picard2001toward,zacharatos2014automatic,chun2016determining}.

\subsection{Classification by Grouping}

In grouping classification, the response of individuals are 
inputs to the machine learning algorithm which 
outputs the group classification.
This is different from classifications of preference and of emotion where the responses are direct outputs of the classification.

Teachers used classification by grouping to assess students on the appropriate level of training to be administered
\cite{aimeur2002clarisse,beck2000high,castillo2003adaptive}.
This gave them an idea of the optimal strategy to be adopted for each group of students, especially when a considerable disparity was observed from the grouping assessment. 
On the other hand, psychologists used group classification to assess mental conditions or capabilities \cite{cetintas2010automatic,karstoft2015early,kersting2014automated}
in order to carry out the further intervention, or to perform an appropriate level of service. Once the group classification was determined, one will only need to match a predefined intervention that was appropriate for the corresponding group.

From the groupings,
a model of classification can be formed. This may be a new grouping with new characteristics or an existing grouping with modified characteristics 
\cite{webb2001machine,razzaghi2016context}.
This is different from the rules classification because 
in this case, the output is groupings and not rules.
As the groupings are formed, the model of the groupings may change. 
Then the characteristics are determined based on the new grouping models, such that the behavior of a group can be predicted.
The last method of grouping classification is very much related to the students (or training) classification but the method of determining the grouping was through gaming
\cite{walonoski2006detection,barata2016early}.
This may result in a more appropriate grouping for younger students because normally they are more alert during a game interaction, which may help in getting a more accurate response from them.

\subsection{Classification by Rules}

Lastly, the classification of rules establishes relationships among different user responses in order to influence a decision-making process. 
It does not necessarily output a final classification but modifies policies or methods that influence the desired output.

One such classification was through open answers to questionnaires \cite{yamanishi2002mining,kersting2014automated}
where classification and association rules were defined 
to characterize targets and establish relationships among them. Although the answers were open, from the keywords and phrase level, their models can be created, and thus the rules that define their relationships can be formed.
Rule classification was also used in determining 
underlying rules to make expert decisions \cite{terano1996knowledge,wheeler2013inside}
even with noisy data.
Rules were classified in determining the casual relationships and word meanings \cite{Tenenbaum2006309,Boose1985495}, in order to understand the idea of what the person is trying to convey. This can be used in spoken language or from documents to develop models for inductive learning and reasoning, or from construct psychology.

Classification of rules for tutoring systems and risk-of-bias assessments were studied in \cite{jarvis2004applying,millard2016machine}. 
In the tutoring system, the purpose was to automate rule generation in the system development, such that
production rules were generated from marked examples and background knowledge. 
In the assessment study, 
model rules were generated for the 
properties of sequence generation, allocation concealment, and blinding. 
The models predicted sentences that contain relevant information, as well as the risk of bias for each research article.
This work has proposed four classifications of the state of mind and emotion
using the different data-gathering methods shown in the previous section.
These classifications enable us to see the 
different aspects of the state of mind and emotion, 
such that its range of forms was discussed.  
To verify these different aspects 
we need to select an experimental platform that enables
us to gather data
from its range of forms in order to
gain a deeper understanding of its nature. 
In this work, we choose addiction as the experimental platform. What is unique to addiction is that 
it covers all the four classifications discussed in this work.

Preference,
which is a choice by 
feelings that involves a mental process or intuition,
is performed when people have manageable addiction. 
It is normally done when decisions are not driven by urges or emotion. On the other hand, choices made by emotion are normally done by people who have a higher degree of addiction. The choices do not anymore involve a mental
process.
In terms of classification by grouping, the choices made by the respondents are analyzed by psychologists that output the groupings.
And lastly, classification by rules that involves the process of looking for new rules or modifying existing ones applies to addiction because when gathering data via questionnaires, the relationships among the questions are 
verified and modified accordingly. This can lead towards understanding the dimensions of addiction and can affect its final output classification.
Thus addiction indeed covers all four classifications discussed in this section.


\section{Machine Learning Classifiers}

Two machine learning classifiers are considered: artificial neural network and support vector machine.
The number of questions will be equal to the dimension of the input space, $n$.
For an $i$-th sample, the corresponding answers can be true ($1$), false ($0$), or 
a degree of state ($[10,100]$). Thus for an input $\mathbf{x}_i\in \mathbb{R}^n$, 
a function $f:\mathbb{R}^n \rightarrow \mathbb{R}$ 
is defined as

\begin{equation}
y_i = f(\mathbf{x}_i)
\end{equation}
where $y_i\in\{1,0,[10,100]\}$. The function $f$ is numerically derived from an artificial neural
network or support vector machine.

\subsection{Artificial Neural Network}

Artificial neural network (ANN) has been extensively used in many different machine learning applications. Two widely used types are feedforward multilayer perceptron and radial basis function.

For multilayer perceptron, given input layer $i$ and output layer $j$, a weight between layers $i$ and $j$ is denoted as $w_{ij}$, 
such that for $n$ nodes in layer $i$,  $\mathbf{w}_j\in\mathbb{R}^n=[w_{1j},\ldots,w_{ij},\ldots,w_{nj}]$. An output of a single node in layer $j$ for a given input $\mathbf{x}\in\mathbb{R}^n$ can be expressed as  

\begin{equation}
\label{eqn:single-layer}
y_j= \sum_{i=1}^n w_{ij} x_i.
\end{equation}

\noindent For an input layer $i$, hidden layers $j$ and $k$, and a single node output (output layer $l$), we can recursively apply (\ref{eqn:single-layer}) three times, to get the input-output relations to be 

\begin{equation}
y_l = \sum_{k=1}^q~w_{kl}~f
\left(
\sum_{j=1}^p~w_{jk}~f
\left(
\sum_{i=1}^n w_{ij} x_i
\right)
\right)
\end{equation}

\noindent such that 
$\mathbf{w}_l\in\mathbb{R}^q$, 
$\mathbf{w}_k\in\mathbb{R}^p$, 
$\mathbf{w}_j\in\mathbb{R}^n$, and
$y=f(\cdot)$ is called the activation function.

For radial basis function with one single output, 
given input $\mathbf{x}$ and number of samples $m$, the following equation can
be applied

\begin{equation}
y = \sum_{j=1}^m w_j~\phi(\|~\mathbf{x}- \mathbf{x}^{(j)}\|)
\end{equation}

\noindent where 
$\phi(\cdot)$ is a set of radial basis functions,
$\mathbf{x}^{(j)}$ is a center of the radial basis function, and 
$w_j$ is an unknown coefficient.

\subsection{Support Vector Machine Model}

Support vector machines (SVM) are derived from statistical learning theory \cite{vapnik1998statistical}.
It has two major advantages over other machine learning tools: (1) it does not have a local
minimum during learning, and (2) its generalization error
does not depend on the dimension of the space. Given $m$ samples $(\mathbf{x}_i,y_i)$
where $i= 1,\ldots,m$, for an
$i$-th sample input $\mathbf{x}_i \in \mathbb{R}^n$,
a scalar offset $b \in \mathbb{R}$ and
a weighting vector $\mathbf{w} \in \mathbb{R}^n $,
a function $f$ is given as

\begin{equation}
f(\mathbf{x}_i) = \mathbf{w}\cdot \mathbf{x}_i + b.
\end{equation}

A loss function $L$ that is insensitive to tolerable error $\epsilon$ can be expressed as 

\begin{equation}
L = \|\mathbf{w}\|^2 + \frac{C}{m}\sum_{i=1}^{m} \max\{0,|y_i - f(\mathbf{x}_i)|- \epsilon \}
\end{equation}
where $C\in \mathbb{R}$ is a regularization constant which can be expressed as an optimization
problem in the form 

\begin{equation}
\begin{split}
\min~~~~
& \frac{1}{2} \| \mathbf{w} \|^2 +  \frac{C}{m}\sum_{i=1}^{m}(\xi_i + \xi_i^*)\\
\mbox{subject to:} ~~~
& ( \mathbf{w} \cdot \mathbf{x}_i +b )- y_i \leq \epsilon + \xi_i\\
& y_i - ( \mathbf{w} \cdot \mathbf{x}_i + b ) \leq \epsilon + \xi_i^* \\
& \xi_i,\xi_i^*\geq 0 \mbox{~~~for~~} i=1,\ldots,m.
\end{split}
\end{equation}

To test the proposed method, an online addiction questionnaire with 10 questions was created 
and answers from 292 respondents were analyzed.
Dependence/independence of questions were verified 
by removing questions
one by one and noting the resulting accuracy of classification,
which can be further developed to 
determine the dimensions of addiction.

\section{Addiction as an Experimental Platform}


To test the proposed machine learning tool, addiction is used as an experimental platform
because it encompasses the entire range of forms of state of mind and emotion, especially based
on its four classifications stated in the previous section.
Depending on the extent of addiction, the person's intuitive response 
can be consciously or unconsciously made. 
When one is not addicted to a stimulant, his 
choices are consciously made and he is in total control of his reaction.
For the state of emotion, normally, the emotional reaction is not consciously controlled but results from an urge or a bodily reaction that automatically occurs given the right stimulant. That is why some people easily cry at sad movies or laugh at certain types of humor. But when one is addicted to a stimulant, the person's reaction is based on an urge or an uncontrolled bodily reaction, similar to the emotional reaction. The person's choice, in this case, is based on unconscious preference, and his reaction is based on bodily urges.
Thus, the study on addiction offers a platform that considers conscious and unconscious decisions, as well as controlled and uncontrolled reactions.
Furthermore, the intervention by a psychologist to come up with groupings based on
inputs from respondents shows classification by grouping.  
And finally, the classification by rules points to the attempt in identifying the dimensions 
of addiction by characterizing the interdependence of inputs from respondents.


\section{Experimental Results}


A questionnaire \cite{jamisola2016surveryquest}
was designed to gather information regarding respondents' degree of addiction to an activity.
The questions were all composed by the author who has no formal training in psychology, thus may be considered as random questions. 
This is done in order to mimic the method of gathering random questions from users that will be included in the database.
These questions do not claim completeness in addressing all the dimensions of addiction
but are presented in order to show how any given randomly gathered set of questions are processed and analyzed.
The analysis is in terms of their interdependence, which may possibly lead towards clustering questions in the database and furthermore, may possibly lead towards identifying the dimensions of addiction. 
At the end of the questionnaire the respondent will rate self as `addicted', `not addicted', `manageable', and `don't know'. 

A total of 
292 
respondents participated in the survey. Ten questions were asked, with possible answers  
`yes', `no', and `not really', as well as a range of numbers to rate the frequency of occurrences or number of persons involved. 
Most of the respondents are students and staff from the University of the author where the average age ranges from 20-30 years old.
Information about the sex of respondents was not gathered because the addiction study in this work is intended to be independent of this information,
including other biases like culture, educational attainment, sexual orientation, race, etc. The author envisioned millions of responses from all over the world that are normalized to any biases due to the randomness of the respondents.
The actual questionnaire and the percentage of responses are shown in Appendix \ref{appendix:questionnaire}.

\subsection{Experimental Setup}

Two machine learning experiments were performed 
using Matlab R2017a neural network, and statistics and machine learning toolboxes. 
The neural network toolbox used `newff' function to create a feed-forward backpropagation network with 
85 hidden layers, four output layers with four outputs that represent four classifications.
Data division was random such that from 292 samples, 204 are used for training, 44 for validation, and 44 for testing. 
The training algorithm used the Levenberg-Marquardt backpropagation technique, while the performance measure was by mean squared error.
After the network has been trained, validated, and tested through `train' command, the network was again tested using `sim' the command that used all the 292 samples, then the output was compared against known target values. 
The average accuracy was at 77\%.



The SVM experiment used `templateSVM' function to create an SVM template that invoked the Gaussian kernel function. 
Then `fitcecoc' was called to
train the classifier using the SVM template that was created. The purpose was to group the responses into four classifications. The training function used 
binary learner and one-versus-one coding design. 
After the SVM training, 10-fold cross-validation by `crossval' command was used.
Then the command `resubPredict' was used to predict the classification from all learners. Its results were compared using the target values from the 292 samples. SVM accuracy was measured at 85\%.

\subsection{Analysis of Answers by Respondents}

In this subsection, we are going to analyze the answers of the respondents based on their own self-assessment of whether or not they are addicted to activity. Of the 292 respondents, $20\%$ considered themselves addicted to the activity, $23\%$ not addicted, $45\%$ manageable, and $12\%$ did not know their status. 
The case of ``manageable'' could mean that the person enjoys the activity but is in total control of his decisions and reactions to it, and is thus not addicted. On the other hand, it can also mean that the person is partially addicted, and has some control over his decisions and reactions to the activity.  Note that an addicted person, as defined above, has totally uncontrolled decisions and reactions.


On the frequency of performing the activity, $50\%$ answered ``everyday.'' However, less than half of this number admitted addiction to the activity.
This result showed that the everyday performance of an activity that one likes to do does not necessarily mean addiction to that activity.
This also means that it is possible for the person to enjoy the addictive activity every day, but is still in control over it.
Regarding the urge to do the activity, $59\%$ admitted to feeling the urge but only one-third of them considered themselves addicted. 
This percentage is higher compared to the percentage of everyday activity, which means
that those who felt the urge, did not necessarily perform the activity every day. 
Furthermore, feeling the bodily urge to do an addictive activity does not easily overcome conscious actions and decisions.

On non-performance of the activity on a regular basis and affecting the mood of the respondent, $43\%$ answered ``yes''.
But only half of this percentage admitted addiction. This is interesting because it means that even without being addicted and in control over the activity, the person can still be affected in his regular daily work through his moods. 
It can also mean that the effort to control the urge to do the activity can somehow affect the everyday mood of the person.
Solitary performance of an addictive activity, with a response of $50\%$, does not necessarily equate to addiction. 
A higher percentage felt the urge to do the activity, but this does not necessarily mean that they are going to do the activity alone.

Having many other major activities besides the addictive activity can be a possible source of getting one's focus away from the addictive activity. 
However, $66\%$ answered the least number of other activities at ``three more'' and $19\%$ answered ``many.''
Thus, in this case, the addicted person can have many other activities besides work and study. 
This can also mean that even the not-addicted person has a limited number of activities besides work and study.


Talking to close friends and family every day as a support group can be vital in coping with addiction. Around $45\%$ talked to ``one or two'' and $36\%$ talked to ``three to five.'' Isolation, in this case, does  not seem to have a close connection to addiction, as in the case of performing the activity alone. 
This could mean that the person can be having many friends and seemed to have a normal life, but is addicted. 
On asking for professional help to stop the activity, a huge percentage of $75\%$ answered ``no.'' This could mean a lack of access to professional help or the hesitation to admit the need for help.

Distraction from daily work or studies caused by the addictive activity has $26\%$ who answered ``yes.'' This is close to the percentage who admitted addiction to the activity. One may say that in this case, a distraction from daily routine caused by the addictive activity is a clear indication of addiction.
(This will be supported by the result in the next section showing this as a critical question.) 
Talking about the activity to somebody else as a possible source of support has $36\%$ answered ``no, I keep it to myself.'' This scenario of isolating oneself is related to the question about the solitary performance of the addictive activity, and to the number of close friends and family that one talks to every day.   
It is noted that the performance of the activity alone has a higher percentage, which means that of all the persons who may be performing the activity alone, a large percentage of it kept it as a secret.

A number of things can be noted in order
to improve the machine learning results. One has to design the questions that tackle independent aspects of the psychological state. This will enhance a clearer separation in the classification. Another possible approach is to create subtle support questions to verify consistency in the answers of the respondent, most especially to critical questions. 
Lastly, indirect questions can be designed so as to avoid the respondent explicitly hide the truthful answer.

\subsection{Investigating the Dimensions of Addiction}

This subsection will analyze 
the dependence/independence of one question
against the rest of the questions based on the output classification accuracy. 
Using the trained model, each question (response) was removed from the input data, and the accuracy of the output was observed. If the accuracy of the output drastically reduced in the absence of a given question, this means the question was critical and was independent of the rest of the questions.
On the other hand, if a given question was removed and the resulting output did not change drastically, that means it was dependent on at least one other question.
In other words, that question did not matter. In this study, a drastic decrease means a $25\%$ reduction from the overall accuracy.

There were 10 questions that each respondent had to answer, and question ten ($Q_{10}$) was a self-assessment based on the four classifications.
Using the model that was created,
responses to question one ($Q_1$) up to question nine ($Q_9$) were removed one by one 
from the input data,
and the accuracy of the output was compared against the target values, that is, the responses to $Q_{10}$. The SVM classification model was used in this analysis.

From the overall accuracy of $85\%$, an accuracy reduction of $25\%$ is an output accuracy of around $64\%$. Table \ref{tab:resultsanalysis} shows the resulting percentage accuracy of classification when
at most two questions were removed. The diagonal elements in the table (in boldface) represent the percentage accuracy when question $Q_i$ was removed. 
(In the table, $i=1,\ldots,9$.)
The encircled values show an accuracy reduction of $25\%$ or more.
The percentage accuracy shown in row $Q_i$ 
is the case when question $Q_i$ was removed first and questions $Q_1$ to $Q_9$ were removed second, one at a time, except $Q_i$. Thus using the convention (row, column) to define the elements in the table, the percentage accuracy in ($Q_1$, $Q_1$) is the case when only question one was removed, and ($Q_1$, $Q_2$) is the case when $Q_1$ was removed first and $Q_2$ was removed second.

\begin{table}[tb!]
\centering
\renewcommand{\arraystretch}{1.4}
\caption{Percentage Accuracy with at Most Two Questions Removed}
\label{tab:resultsanalysis}
\begin{tabular}{c|c|c|c|c|c|c|c|c|c}
\centering
$Q_i$ & $Q_1$ & $Q_2$ & $Q_3$ & $Q_4$ & $Q_5$ & $Q_6$ & $Q_7$ & $Q_8$ & $Q_9$  \\ 
\hline \T\B
$Q_1$ & \textbf{81} & 76 & 80 & 74 & 78 & 66 & 79 & $\circled{49}$  & 70 \\
\hline\T\B 
$Q_2$ & 77 & \textbf{82} & 74 & $\circled{60}$ & 80 & 67 & 78 & $\circled{58}$  & 78 \\
\hline\T\B 
$Q_3$ & 80 & 74 & \textbf{82} & 76 & 77 & 74 & 75 & $\circled{59}$  & 73 \\
\hline\T\B 
$Q_4$ & 70 & $\circled{60}$ & 76 & \textbf{79} & 78 & 74 & 80 & $\circled{61}$  & 72 \\
\hline\T\B 
$Q_5$ & 80 & 80 & 78 & 79 & \textbf{83} & 76 & 81 & 73  & 74 \\
\hline\T\B 
$Q_6$ & 66 & 68 & 76 & 75 & 75 & \textbf{78} & 74 & $\circled{57}$ & 66 \\
\hline\T\B 
$Q_7$ & 79 & 78 & 75 & 79 & 81 & 68 & \textbf{81} & $\circled{55}$ & 81 \\
\hline\T\B 
$Q_8$ & $\circled{62}$ & 71 & $\circled{59}$ & $\circled{60}$ & 76 & $\circled{58}$ & $\circled{58}$ & $\circled{\mbox{\textbf{45}}}$ & 65 \\
\hline\T\B 
$Q_9$ & 74 & 77 & 73 & 73 & 74 & 70 & 78 & $\circled{60}$  & \textbf{80} \\
\hline

\end{tabular}
\end{table}

From Table \ref{tab:resultsanalysis}, it can be observed that removal of $Q_8$ drastically lowers the output 
classification accuracy, which resulted in $(Q_8,Q_8)= 45\%$. This 
drastic decrease in accuracy 
is generally consistent all throughout the elements of $Q_8$ row and column, except for $(Q_5,Q_8)=75\%$ and 
$(Q_8,Q_5)=76\%$. 
This could mean that for most elements in $Q_8$ row and column, the removal of $Q_8$ and one other question, is greatly influenced by the absence of $Q_8$ alone. The other question did not greatly affect the accuracy results, except $Q_5$. We note question eight below.

\bigskip

\noindent \texttt{$Q_8$: Do you think you get distracted in your daily work or studies by thinking about this activity?}

\bigskip


\noindent This could mean that distraction from the daily activity is generally independent of the rest of the questions in the addiction survey questionnaire. Thus, it can be considered a critical question and can be counted as an independent dimension of addiction.

Removal of $Q_2$ and $Q_4$ resulted in a more drastic decrease in accuracy compared to the removal of $Q_2$ or $Q_4$ alone. This could mean that both $Q_2$ and $Q_4$ belong to one class of critical questions, which are independent of the rest of the questions. 
\bigskip

\noindent \texttt{$Q_2$: Do you feel an urge to do it?}

\noindent \texttt{$Q_4$: Do you do this activity alone or with some company?}

\bigskip

\noindent The relationship can be that the feeling of a strong urge to do an addictive activity is somehow related to doing such activity on one's own accord, that is, being alone. And when the urge is lesser, it is somehow related to the performance of the activity with more company. Thus, we can say that another independent dimension of addiction includes the urge of performance or the number of persons involved during the performance.

A peculiar observation is $Q_5$ and $Q_8$ and we note below. Question $Q_5$ is stated in the following.

\bigskip

\noindent \texttt{$Q_5$: Besides work or studies, how many other main activities you have in a day besides this activity?}

\bigskip
We note that the removal of both $Q_5$ and $Q_8$ resulted in a higher accuracy compared to the removal of $Q_8$ alone, such that
$(Q_8,Q_5)=76\%$ and $(Q_5,Q_8)=73\%$.
That is, the removal of both $Q_5$ and $Q_8$ resulted in a $10\%$ accuracy reduction against the overall accuracy, but removal of $Q_8$ alone results in a $25\%$ accuracy reduction.
We note further that the removal of $Q_5$ alone had almost zero percent accuracy reduction, and is in fact the highest accuracy that is closest to the overall accuracy. 
When $Q_8$ was removed, the new accuracy was drastically reduced to $45\%$. But when $Q_5$ was removed next, the new accuracy drastically increased to $76\%$. This means that $Q_5$ was dependent on $Q_8$ alone such that when $Q_8$ was removed, it became an independent question and did affect the accuracy drastically.
Let's investigate now the reverse order of removal. When $Q_5$ was removed, the accuracy did not change much and was at its highest value among the rest of the single questions removed. But when $Q_8$ was removed, the accuracy did not drastically change. This means that the removal of $Q_5$ affected $Q_8$ such that it was not able to drastically change the accuracy as it did when the other questions were removed. Thus, $Q_8$ was dependent on $Q_5$. But initially, we identified $Q_8$ to be a critical question because it drastically changed the accuracy when removed alone. The explanation is that the characteristic of $Q_5$ was very similar to the overall accuracy such that when $Q_8$ was removed alone, this dependence was not obvious.

Another observation is the relationship between $Q_8$ and $Q_2$. The case of $(Q_2,Q_8)=59\%$ but $(Q_8,Q_2)=71\%$, that is, the order of removal has an effect on the resulting accuracy.
In the first case, the accuracy did not drastically change when $Q_2$ was removed. The drastic change of $59\%$ happened only when $Q_8$ was removed after $Q_2$. This is the same case as when $Q_8$ was paired with the rest of the questions, except $Q_5$. Which means that the removal of $Q_2$ did not affect the removal of $Q_8$, and therefore $Q_8$ is independent of $Q_2$.
In the second case, 
removal of $Q_8$ resulted in a drastic decrease of accuracy to $45\%$, but when $Q_2$ was removed after $Q_8$, the accuracy drastically increased to $71\%$. Thus $Q_2$ was affected by the removal of $Q_8$ and is therefore dependent on $Q_8$. But $Q_8$ is not dependent on $Q_2$, thus the dependence is only in one direction and not both. This explains why the order of removal has an effect on the resulting accuracy.

The approach presented above can be used to identify critical questions that drastically change the accuracy output when removed. 
Critical questions can help identify the number of independent variables
in the state of mind and emotion and can help in determining its dimensionality.
Identifying critical questions can also help in minimizing the questions asked in the questionnaire, in order to save time for the respondents. 
This possibility of quantifying the dimensionality of a person's state of mind and emotion by an individual with no sufficient background in psychology upholds the advantage of a machine learning tool that can help replace the  ``expertise'' required to perform an intelligent evaluation. 
It is noted that the identification 
of this dimensionality 
can be very difficult for an experienced psychologist to discover.

Another future direction of the proposed method is the possibility of developing questions with hidden information such that the respondent cannot intentionally cheat on his 
answers. 
In addition, questions or choices of answers can be designed to capture a faster response from the respondents, such that the questionnaire can be more user-friendly. This way, user-friendliness from the perspective of the respondent can be accommodated without compromising technicality from the psychologist's perspective.
Furthermore, this can lead to drastically
increasing the number of questions in the database, such that the addicted person can test himself 
again without answering exactly the same set of questions.
This can make the self-assessment more reliable.

\section{Survey Questions vs. SADD Questionnaire}

A questionnaire designed by the author to assess the degree of addiction to an activity by a respondent, called ``A Survey on Addiction,'' is shown in Appendix \ref{appendix:questionnaire}. 
This set of questions will be  compared against a standard psychological test called 
``Short Alcohol Dependence Data (SADD)'' Questionnaire
\cite{raistrick1983development} that is used to assess oneself to alcohol addiction. And secondly, we will analyze the answers of the 292 respondents that participated in the 
addiction survey. 


One major difference in the addiction survey questions in this paper compared to the SADD questions are that the questions in this paper assessed addiction before the tangible effects are experienced. 
They did not tackle cases about physical effects of addiction like ``shaky hands'', ``vomiting'', ``imaginary'' things, etc. but these are included in the SADD questions. 
In the following discussion, we compare the first few questions from SADD against the questions of the survey shown in Appendix \ref{appendix:questionnaire}.

Question one of SADD addresses the issue of getting the thought of drinking out of the mind, and this is similar to question eight in the survey which asked regarding the thought on the addictive activity being distractive to daily work or studies. 
Obviously, when something distracts your daily routine, it means the thought about it is always in the mind.

Question two of SADD talked about misplaced priorities due to alcohol addiction. This is related to question five in this work that talked about major activities including addictive activity. But the SADD question is transparent in asking about misplaced priorities. Being transparent in the question can be an advantage to get a clear answer regarding it. Or this can be a disadvantage as well when the respondent will try to suppress from giving an accurate answer. Thus an indirect question might be able to address this issue.

Question three of SADD where the activities of the respondent are revolving around alcohol drinking is again related to question five that asked about major activities of the respondent including the addictive activity.
That is, if the addictive activity constitutes a major
activity of the respondent then the addictive activity greatly influences all the other activities.
Question four of SADD considers the frequency of drinking alcohol and is related to question one in this work that explicitly asks about the frequency of performing the addictive activity. 
In this question, the two approaches are very closely related. 

Question five of SADD asked about the desire to satisfy the need for alcohol without considering the quality of the drink and is related to question two of this work that asked about the urge to perform the addictive activity. In this way, the urge to do the addictive activity created the possibility to disregard any discomfort that may be experienced in performing it.
Lastly, there were questions in SADD asked about the awareness of possible consequences of drinking alcohol. This is related to question two of this work that considered the urge to do the addictive activity without considerations of possible consequences. It can also be related to question eight that asks about seeking professional help because of the possible consequences of the addictive activity.

\section{Conclusion and Future Direction}
This paper has shown the possibility of determining the state of mind and emotion of an individual through a questionnaire-based machine learning tool, using an artificial neural network and support vector machine. 
Previous classifications and data-gathering methods were presented to determine preference, opinion, emotion, or capability. 
The proposed method is implemented in analyzing addiction through a survey on addiction with ten questions. 
Results analysis showed a proposed method to identify critical questions that can lead to the identification of the dimension of addiction.
Analysis of the survey questions against a standard questionnaire on alcohol addiction is presented. 
This tool can be used to do the same method of computation for all applications but will vary only on the types of questions asked depending on the individual information to be extracted. The proposed machine learning diagnostic tool may be able to output judgment,
based on the thousands of inputs collected from users.
{
The future direction of this research is for
a psychologist to assess, compare and validate the 
proposed method and its results. 
In addition, a deeper investigation of the dimensions of 
addiction via a machine learning model will be performed.
Lastly, as the online database of questions and answers increased, 
it is recommended to use unsupervised machine learning to build the state of mind and emotion model
through the correlation of the responses from the respondents.
}


\section*{Acknowledgment}

The author would like to acknowledge the contribution of 
Mario Saiano,
Social Health Educator,
Local Health Unit Genovese, Italy 
for his inputs in the
preparation of this manuscript.


\section{The Questionnaire on Addiction Survey}
\label{appendix:questionnaire}

A survey is designed to assess the addiction of a respondent to an activity. The survey was posted online using Google forms \cite{jamisola2016surveryquest}. This section shows the instruction, questions, and responses from 292 respondents. 

\bigskip

Instruction: This survey consists of 10 questions. Please be honest in answering. Think of one type of activity that you like, answer the following questions, and judge for yourself at the end of the survey if you consider yourself addicted or not to this activity.

\begin{table}[tb!]
\centering
\label{tab:survey-questions}
\caption[Caption for LOF]%
{A Survey on Addiction}
\begin{tabular}{ll}
\hline
\multicolumn{2}{l} 
{1. How often do you do this activity?} \T\B \\
$\bullet$ Everyday (50.3\%) & $\bullet$ Twice a day (4.5\%)	\\
$\bullet$ Once a week (24\%)       & $\bullet$ Twice a week (21.2\%)	\\

\hline
\multicolumn{2}{l}{2. Do you feel an urge to do it? } \T\B\\
$\bullet$ Yes (59.2\%) & $\bullet$ No (8.6\%) 	\\
$\bullet$ Not really (32.2\%)       &  \\

\hline
\multicolumn{2}{l}{3. Does it affect your mood if you do not  do this activity on} \T\B \\
\multicolumn{2}{l}{ a regular basis?} \\
$\bullet$ Yes (43.2\%) & $\bullet$ No (42.8\%) 	\\
$\bullet$ Not sure (14\%)       &  \\

\hline
\multicolumn{2}{l}{4. Do you do this activity alone or with some company? } \T\B\\
$\bullet$ Alone (50.3\%) & $\bullet$ Two to three (23.3\%) 	\\
$\bullet$ More than three (3.1\%)       &  $\bullet$  Does not matter (23.3\%)\\

\hline
\multicolumn{2}{l}{5. Besides work or studies, how  many other main activities you have} \T\B\\ 
\multicolumn{2}{l}{in a day besides this activity?}  \\
$\bullet$ Three more (66.1\%) & $\bullet$ Five more (14\%) \\
$\bullet$  10 more (1.4\%) &  $\bullet$  Many (18.5\%)	\\

\hline
\multicolumn{2}{l}{6. How many very close friends and family do you talk to everyday?} \T\B\\
$\bullet$ One or two (44.5\%) & $\bullet$ Three to five (36\%) \\
$\bullet$  Around 10 (8.2\%) &	 $\bullet$ Many (11.3\%)\\

\hline
\multicolumn{2}{l}{7. Did you attempt to seek professional help to stop this activity?} \T\B \\
$\bullet$ No (74.7\%) & $\bullet$ Yes (6.8\%) \\
$\bullet$  Not really (18.5\%) &	 \\

\hline
\multicolumn{2}{l}{8. Do you think you get distracted in your daily work or studies by} \T\B \\
\multicolumn{2}{l}{ thinking about this activity?}\\
$\bullet$ Yes (25.7\%) & $\bullet$ No (48.3\%) \\
$\bullet$  Manageable (26\%) &	 \\

\hline
\multicolumn{2}{l}{9. Have you talked with others  about this activity? }\T\B \\
\multicolumn{2}{l}{$\bullet$ No, I keep it to myself (36.3\%)} \\
\multicolumn{2}{l}{$\bullet$ Selected few (32.2\%)}\\
\multicolumn{2}{l}{$\bullet$ Close friends and family only (15.4\%)}\\
\multicolumn{2}{l}{$\bullet$ Everybody knows about it (16.1\%)} \\

\hline
\multicolumn{2}{l}{10. Rate yourself with regard to  this activity} \T\B \\
$\bullet$ Addicted (19.5\%) & $\bullet$ Not addicted (23.3\%) 	\\
$\bullet$ Manageable (44.9\%)       &  $\bullet$  I don't know (12.3\%)\\

\hline
\end{tabular}
\end{table}

\footnotesize
\bibliographystyle{IEEEtran}
\bibliography{IEEEfull,jamisolamachinelearning}

\end{document}